\def\year{2018}\relax
\documentclass[letterpaper]{article} 
\usepackage{aaai18}  
\usepackage{times}  
\usepackage{helvet}  
\usepackage{courier}  
\usepackage{url}  
\usepackage{graphicx}  
\usepackage{booktabs}       
\usepackage{amsfonts}       
\usepackage{nicefrac}       
\usepackage{microtype}      

\usepackage{graphicx}
\usepackage{algorithm} 
\usepackage{algorithmic}
\usepackage{amsthm}
\usepackage{amssymb}
\usepackage{multicol}
\usepackage{amsmath}

\title{Hyperparameter Tuning for the Contextual Bandit}
\author{Djallel Bouneffouf$^1$, Emmanuelle Claeys$^2$\\
$^{1}$IBM Thomas J. Watson Research Center, Yorktown Heights, NY USA\\
$^2$  Strasbourg University, CNRS, IRMA, 7 Rue Rene Descartes, 67000 Strasbourg, France \\
\{Djallel.bouneffouf\}@ibm.com \\
claeys@unistra.fr
}
\begin{document}

\maketitle

\begin{abstract}
We study here the problem of learning the exploration exploitation trad-off in the contextual bandit problem with linear reward function setting. In the traditional algorithms that solve the contextual bandit problem, the exploration is a parameter that is tuned by the user. However, our proposed algorithm learn to choose the right exploration parameters in an online manner based on the observed context, and the immediate reward received for the chosen action. We have presented here two algorithms that uses a bandit to find the optimal exploration of the contextual bandit algorithm, which we hope is the first step toward the automation of the  multi-armed bandit algorithm. 
\end{abstract}

\section{Introduction}

In sequential decision problems such as clinical trials \cite{villar2015multi} or recommender system \cite{MaryGP15,BouneffoufBG13}, a decision-making algorithm must select among several actions at each given time-point. Each of these actions is associated with side information, or context (e.g., a user's profile), and the reward feedback is limited to the chosen option. For example, in the clinical trials \cite{villar2015multi,BouneffoufRC17,BouneffoufBG12}, the actions correspond the treatment options being compared, the context is the patient's medical record (e.g. health condition, family history, etc.) and the reward represents  the outcome (successful or not) of the  proposed treatment. In this setting, we are looking for  a good trade-off between the exploration of the new drug and the exploitation of the known drug.

This inherent exploration-exploitation trade-off exists in many sequential decision problems, and is traditionally modeled as {\em multi-armed bandit (MAB)} problem, stated as follows: there are $K$ ``arms'' (possible actions), each associated with a fixed but unknown  reward probability distribution ~\cite {LR85,UCB,dj10}. At each step, an agent plays an arm (chooses an action) and receives a reward. This reward is drawn according to the selected arm's law and is independent of the previous actions.

A particularly useful version of MAB is the contextual multi-armed bandit problem. In this problem, at each iteration,  before choosing an arm, the agent observes a $D$-dimensional {\em feature vector}, or {\em context}, associated with each arm. The learner uses these contexts, along with the rewards of the arms played in the past, to choose which arm to play in the current iteration. Over time, the learner's aim is to collect enough information about the relationship between  the context vectors and rewards, so that it can predict the next best arm to play by looking at the corresponding contexts (feature vectors) \cite{AgrawalG13}.
One smart solution for the contextual bandit is the LINUCB algorithm, which is based on online ridge regression, and takes the concept of upper-confidence bound \cite{Li2010} to strategically balance between exploration and exploitation. 
The $\alpha$ parameter essentially controls exploration/ exploitation. The problem is that it is difficult to decide in advance the optimal value of $\alpha$ We introduce in this paper two algorithms, named "OPLINUCB" and " DOPLINUCB", that computes the optimal value of $\alpha$ in both stationer and switching environment by adaptively balancing exr/exp according to the context.  

The main contributions of this paper include proposing two new algorithms, for both stationary and non-stationary settings, which extend the existing bandit algorithms to the new setting, and (3) evaluating the algorithms empirically on a variety of datasets.

\section{RELATED WORK}
\label{sec:related}
The multi-armed bandit problem is a model of exploration versus exploitation trade-off, where a player gets to pick within a finite set of decisions the one   maximizing the cumulative reward.
This problem has been extensively studied. Optimal solutions have been provided using a stochastic formulation ~\cite{LR85,UCB,Bouneffouf0SW19,LinBCR18,BF16,DB2019,BalakrishnanBMR19ibm,BouneffoufLUFA14,RLbd2018}., a Bayesian formulation ~\cite {surveyDB,T33,BRCF17}, or using an adversarial formulation ~\cite{AuerC98,AuerCFS02,BalakrishnanBMR18}. However, these approaches do not take into account the context which may affect to the arm's performance.
In LINUCB ~\cite{Li2010,ChuLRS11} and in Contextual Thompson Sampling (CTS)~\cite{AgrawalG13} and neural bandit \cite{AllesiardoFB14}, the authors assume a linear dependency between the expected reward of an action and its context; the representation space is modeled using a set of linear predictors. However, the exploration magnitude on these algorithms need to be given by the user. Authors in \cite{sharaf2019meta}, addresses the exploration trade-off by learning a good exploration strategy for offline tasks based on synthetic data, on which it can simulate the contextual bandit setting. Based on these simulations, the proposed algorithm uses an imitation learning strategy to learn a good exploration policy that can then be applied to true contextual bandit tasks at test time. the authors compare the algorithm to seven strong baseline contextual bandit algorithms on a set of three hundred real-world datasets, on which it outperforms alternatives in most settings, especially when differences in rewards are large. 

In \cite{bastani2017mostly,Bouneffouf16}, the authors show that greedy algorithms that exploit current estimates without any exploration may be sub-optimal in general. However, exploration-free greedy algorithms are desirable in practical settings where exploration may be costly or unethical (e.g., clinical trials). They find that a simple greedy algorithm can be rate-optimal if there is sufficient randomness in the observed contexts. They prove that this is always the case for a two-armed bandit under a general class of context distributions that satisfy a condition they term covariate diversity. Furthermore, even absent this condition, we show that a greedy algorithm can be rate optimal with positive probability.

As we can see, none of these previously proposed approaches involves learning dynamically the exploration trade-off in the contextual bandit setting, which is  the main focus of this work.

\section{Key Notion}
This section focuses on introducing the key notions used in this paper.

\subsection{\bf The Contextual Bandit Problem.}
Following \cite{langford2008epoch}, this problem is defined as follows.
At each time point (iteration) $t \in \{1,...,T\}$, a player is presented with a {\em context} ({\em feature vector}) $\textbf{c}(t) \in \mathbf{R}^N$
  before choosing an arm $i  \in A = \{ 1,...,K\} $.
We will denote by
  $C=\{C_1,...,C_N\}$  the set of features (variables) defining the context.
Let ${\bf r} = (r_{1}(t),...,$ $r_{K}(t))$ denote a reward vector, where $r_i(t) \in [0,1]$ is a reward at time $t$ associated with the arm $i\in A$.
Herein, we will primarily focus on the Bernoulli bandit with binary reward, i.e. $r_i(t) \in \{0,1\}$.
Let $\pi: C \rightarrow A$ denote a policy. Also, $D_{c,r}$ denotes a joint distribution  $({\bf c},{\bf r})$.
We will assume that the expected reward is a linear function of the context, i.e.
$E[r_k(t)|\textbf{c}(t)] $ $= \mu_k^T \textbf{c}(t)$,
where $\mu_k$ is an unknown weight vector (to be learned from the data) associated with the arm $k$. 

\subsection{\bf Thompson Sampling (TS).}
The TS \cite{thompson1933likelihood}, also known as
Basyesian posterior sampling, is a classical approach to multi-arm bandit problem, where the  reward $r_{k}(t)$ for choosing an arm $k$ at time $t$ is assumed to follow a distribution $Pr(r_{t}|\tilde{\mu})$ with the  parameter $\tilde{\mu}$. Given a prior $Pr(\tilde{\mu})$ on these parameters, their posterior distribution   is given by the Bayes rule, $Pr(\tilde{\mu}|r_{t}) \propto Pr(r_{t}|\tilde{\mu}) Pr(\tilde{\mu})$. A particular case of the Thomson Sampling approach assumes a  Bernoulli bandit problem, with rewards being 0 or 1, and the parameters following the Beta prior.
TS initially assumes arm $k$ to have prior $Beta(1, 1)$ on $\mu_k$ (the probability of success). At time $t$, having observed $S_k(t)$ successes (reward = 1) and $F_k(t)$ failures (reward = 0), the algorithm updates the distribution on $\mu_k$ as $Beta(S_k(t), F_k(t))$. The algorithm then generates independent samples $\theta_k(t)$ from these posterior distributions of the $\mu_k$, and selects the arm with the largest sample value. For more details,  see, for example, \cite{AgrawalG12}. \\

\subsection{Regression Tree}
In total generality unsupervised learning is an ill posed problem. Nevertheless in some situations it is possible to back the clustering problem by a supervised one. In our setting we can use the reward estimation as a supervision for the group creation. Thus we just need a supervised learning technique which explicitly creates groups that we could reuse. See  \cite{Kotsiantis07supervisedmachine} for a survey on such techniques. Among all existing approaches, the use of a tree built by recursive  partitioning is a popular approach.  It allows to estimate different means under specifics explanatory variables and have an  interpretation as a regression analysis \cite{CIS-158589}. Moreover some efficient implementations as  "CART" \cite{reason:BreFriOlsSto84a} and "C4.5" are available.

For a regression models describing the conditional distribution of a response variable $X$ given the status of $N_F$ covariates by means of tree-structured recursive partitioning. The $d$-dimensional covariate vector $f_{X}=(f^1_X,..., f^m_X)$ is taken from a sample space $\mathcal{F}$. Both response variables
and covariates may be measured at arbitrary scale. If we assume that the conditional distribution $D(X|f_{X})$ of the respond $X$ variable given the covariate $f_{X}$ depend of a function $z$ of the covariate:
\[
D(X|f_{X}) = D(X|f^1_X,..., f^m_X) = D(X|z(f^1_X,..., f^m_X))
\]
With $m \leq N_F$.
Let's $\mathcal{L}$ be a learning sample for training the model of regression relationship on a $n$ random sample of independent and identically distributed observation, possibly with some covariates observation is missing :
\[
\mathcal{L}_n = \{ X_i,f^1_i,...,f^m_i ; i=1,...,n\}
\]

\texttt{CTree} (conditional inference Tree) is an algorithm derived from C.A.R.T : \texttt{CTree} proposed by \citeauthor{doi:10.1198/106186006X133933} (2006) which is a non parametric class of regression trees embedding tree-structured regression models into a well-defined theory of conditional inference procedures. The main advantage of \texttt{CTree} is to handle any kind of regresssion problemes, including nominal, ordinal, numeric, censored as well multivariate response variable and arbitrary measurement scales of the covariates .  
CTree also manages the bias induced by maximizing a splitting criterion over all possible splits simultaneously (\citeauthor{SHIH2004457}).
The following algorithm consider for a learning sample $\mathcal{L}_n$ as a non negative integer valued case weights $w=(w_1,...,w_n)$. Each node of a tree is represented by a vector of case weights having non-zero elements when the corresponding observations are elements of the node and are zero otherwise. 
\begin{algorithm}
\caption{CTree}\label{CTree}
\begin{algorithmic}[1]
\STATE {\bfseries Require:} A learning sample of data $\mathcal{L}_n$
\FOR{$w$=$w_1$,...,$w_N$}
\STATE For case weights $w$ test the global null hypothesis of independence between $m$ covariates and the response. Stop if hypothesis cannot be rejected. Otherwise select the covariate $\mathcal{F}_j^*$ with the strongest association to $X$.
\STATE Choose a set $b^* \subset \mathcal{B}_j^*$ in order to split $\mathcal{B}_j^*$ into two disjoint sets : $b^*$ and $\overline{b^*}$. Change the case weights for determine the two subgoups (left and right).
\STATE Modify case weight of left and right subset respectively.
\ENDFOR
\end{algorithmic}
\end{algorithm}
This algorithm stops when the global null hypotheses of independence between the response and any $m$ covariates cannot be rejected at a pre-specified normial $\alpha$. 
Details about variable selection, stopping criteria, splitting criteria or missing values and surrogate splits can be found at \citeauthor{doi:10.1198/106186006X133933} (2006). 
This algorithm handles the missing values and uses a Bonferroni correction to counteract the problem of multiple comparisons.
The computational complexity of the variable selection depend of the covariates nature: for continuous variable, searching the optimal split is of order $n\log{n}$ , for nominal covariates measured of $L$ levels , the evaluation of all possible splits is maximize by $2^{L-1}-1$. 


\section{Algorithms for Learning the Exploration Value }
We describe here two algorithm that learn the exploration of the contextual bandit algorithm.

\subsection{OPLINUCB}
The proposed algorithm is named "OPLINUCB"  for Non-Parametric LINUCB. This algorithm has to solve two levels multi-armed bandit problems. The first level is the classical multi-armed bandit problem applied to find the parameters of the algorithm. The second level problem is a contextual bandit problem that use the parameters find in the first level to find the optimal arm to play. 
Let $n_{\alpha_i}(t)$ be the number of times the $i$-th exploration value has been selected so far, let  $r^f_{\alpha_i}(t)$ be the cumulative reward associated with the exploration value $i$, and let $r_k(t)$ be the reward associated with  the arm $k$ at time $t$.
The algorithm takes as an input the candidate values for $\alpha$, as well as the initial values of the Beta distribution parameters in TS.
At each iteration $t$, we update the values of those parameters, $S_{\alpha_i}$ and $F_{\alpha_i}$ (steps 5 and 6), to represent the current total number of successes and failures, respectively, and then sample the ''probability of success'' parameter $\theta_{\alpha_i}$
from the corresponding $Beta$ distribution, separately for each exploration value $\alpha_i$ to estimate $\mu^i$, which is the mean reward conditioned to the use of the variable $i: \mu^i = \frac{1}{K} \sum_k E[ r_k . 1\{i \in C^d \}]$ (step 7).
\\
The pseudo-code of OPLINUCB is sketched in Algorithm~\ref{alg:LINUCB}.
\begin{algorithm}
\caption{ OPLINUCB}
 \label{alg:NLINUCB}
 \begin{algorithmic}
   \STATE {\bfseries Input:} $(\alpha_1, ..., \alpha_N):$ candidate values for $\alpha$, $A_t$,  the initial values $S_{\alpha_i}(0)$ and $F_{\alpha_i}(0)$ of the Beta distribution parameters. 
   \FOR{t=1 {\bfseries to} T} 
   \STATE Observe the context $x_{t} \in R^d$
   \FOR{all $\alpha_i$}
   \STATE $S_{\alpha_i}(t)=S_{\alpha_i}(0)+r^f_{\alpha_i}(t-1)$ 
   \STATE $F_{\alpha_i}(t)=F_{\alpha_i}(0)+n_{\alpha_i}(t-1)-r^f_{\alpha_i}(t-1)$
   \STATE Sample $\theta_{\alpha_i}$ from $ Beta(S_{\alpha_i}(t), F_{\alpha_i}(t))$ distribution
    \ENDFOR
   \STATE Choose exploration $\alpha_t = argmax_{\alpha \in A} p_{t,a} $ 
   \FOR{all $a \in A_t$}
    \IF{ a is new}
   \STATE \textbf{A}$_a \leftarrow I_d $ (d-dimensional identity matrix)
   \STATE $b_a \leftarrow 0_{d*1}$ (d-dimensional zero vector)
    \ENDIF
   \STATE $\Theta_a \leftarrow$ \textbf{A}$_{a}^{-1}*b_a$  
   \STATE $p_{t,a} \leftarrow \Theta^{\top}_a x_{t,a} +
   \alpha_t \sqrt{x^{\top}_{t,a}
   \textbf{A}_{a}^{-1} x_{t,a}}$
    \ENDFOR
   \STATE Choose arm $a_t = argmax_{a\in A_t} p_{t,a} $ with ties broken arbitrarily,
   and observe a real-valued payoff $r_t$
   \STATE \textbf{A}$_{a_t} \leftarrow$ \textbf{A}$_{a_t} + x_{t,a_t} x^{\top}_{t,a_t}$
   \STATE $b_{a_t} \leftarrow b_{a_t} + r_t x_{t,a_t}$
   \ENDFOR
   \end{algorithmic}
\end{algorithm}

In Algorithm~\ref{alg:NLINUCB}, $A_t$ is the set of arms at iteration $t$, where $x_{a,t}$ is the feature vector of arms $a$ with $d$-dimension, $\Theta_a$ is the unknown coefficient vector of the feature $x_{a,t}$, $\alpha$ is a
constant and \textbf{A}$_a=D_{a}^{\top}D_a +I_d$.
$D_a$ is a design matrix of dimension $m \times d$ at trial $t$, whose rows correspond to $m$ training inputs (e.g., $m$ contexts that are observed previously for arm $a$), and $b_a \in R^m$ is the corresponding response vector (e.g., the corresponding $m$ click/no-click user feedback). Applying ridge regression to the training data $(D_a, c_a)$ gives an estimate of the coefficients: $\Theta_a = (D_{a}^{\top} D_a + I_d)^{-1} D_{a}^{\top}c_a$, where $I_d$ is the $d \times d$ identity matrix and $c_a$ are independent conditioned by corresponding rows in $D_a$.
\subsection{DOPLINUCB}
The proposed algorithm is named "DOPLINUCB" for Dynamic Non-Parametric LINUCB. This algorithm has to solve two levels contextual bandit problems. The first level where the algorithm is using the context to decide on the exploration value of the algorithm, this is done using the CTree algorithm described in Algorithm 1. The second level problem is also contextual bandit problem, where the algorithm uses the context and the exploration provided by the first level to find the optimal arm to play. The pseudo-code of DOPLINUCB is sketched in Algorithm~\ref{alg:LINUCB}.
\begin{algorithm}
\caption{DOPLINUCB}
 \label{alg:LINUCB}
 \begin{algorithmic}
   \STATE {\bfseries Input:} $(\alpha_1, ..., \alpha_N):$ candidate values for $\alpha$, $A_t$ 
   \FOR{t=1 {\bfseries to} T} 
   \STATE Observe the context $x_{t} \in R^d$
   \FOR{all $\alpha_i$}
   \STATE $p_{t,\alpha} \leftarrow CTree (\alpha_1, ..., \alpha_N, x_t)$
   \ENDFOR
   \STATE Choose exploration $\alpha_t = argmax_{\alpha \in A} p_{t,a} $ 
   \FOR{all $a \in A_t$}
    \IF{ a is new}
   \STATE \textbf{A}$_a \leftarrow I_d $ (d-dimensional identity matrix)
   \STATE $b_a \leftarrow 0_{d*1}$ (d-dimensional zero vector)
    \ENDIF
   \STATE $\Theta_a \leftarrow$ \textbf{A}$_{a}^{-1}*b_a$  
   \STATE $p_{t,a} \leftarrow \Theta^{\top}_a x_{t,a} +
   \alpha_t \sqrt{x^{\top}_{t,a}
   \textbf{A}_{a}^{-1} x_{t,a}}$
    \ENDFOR
   \STATE Choose arm $a_t = argmax_{a\in A_t} p_{t,a} $ with ties broken arbitrarily,
   and observe a real-valued payoff $r_t$
   \STATE \textbf{A}$_{a_t} \leftarrow$ \textbf{A}$_{a_t} + x_{t,a_t} x^{\top}_{t,a_t}$
   \STATE $b_{a_t} \leftarrow b_{a_t} + r_t x_{t,a_t}$
   \ENDFOR
   \end{algorithmic}
\end{algorithm}

\section{Experimentation}
We show in the following two senarios, one where we evaluate the algorithms in stationary environment, and the second with a non-stationary environment. 
\subsection{Stationary Environment}
In this scenario we consider the case of two Bernouilli amrs,  our dataset is \textit{Adult dataset}\cite{Adult:1994}. 
Reward is a prediction task to determine whether a person makes over 50K a year. 
Each person is define by some categorical and continuous information (age, work class, etc). 
We fix interval of possible $\alpha$ in $[0.01;1]$ with step of $0.01$. Mean , median, min and max of empirical average cumulative of each LinUCB with different $\alpha$ are provided in Table 1. Rewards function of each arm are stationary. We ran LinUCB with a hundred different values of $\alpha$ and compare empirical average cumulative regret with DOPLINUCB and OPLINUCB. Without train set, it's not possible to use DOPLINUCB. Nevertheless, OPLINUCB choose an $\alpha$ in 100 value of alpha. \textit{DOPLINUCB} and \textit{OPLINUCB} can't beats the best $\alpha$ value (min on table \ref{tab:exp_1}) but \textit{OPLINUCB} beats every time the median/mean.

\begin{table}[H]
\centering

\label{tab:exp_1}
\begin{tabular}{l|llll}
size of train set & 0      & 1000    & 5000    & 10 000   \\ \hline
max               & 5216   & 5116    & 4406    & 3658    \\
min               & \textbf{5096}   & \textbf{4964}    & \textbf{4297}    & \textbf{3560}    \\
mean              & 5152.1 & 5035,17 & 5035,17 & 3605,63 \\
median            & 5149   & 5031    & 5031    & 3605    \\
OPLINUCB      & \textbf{5121}   & \textbf{4981}    & \textbf{4334}    & \textbf{3571}    \\
DOPLINUCB   &        & 5014    & 4399    & 3654   

\end{tabular}
\caption{Performance measures for $\alpha \in [0.01,1]$ }
\end{table}

\subsection{Non-Stationary Environment}
In this experiment, we consider a challenging setting: reward function of each arm changes at a fixed number of iteration. 
We keep a training-set of 5000 items. Table 2 show cumulative regrets when reward function of each arm changes at 100/1000/10000 iteration. 

We can observe that in this scenario the proposed algorithm outperform beats the best $\alpha$ value (min on table). This is explainable by the fact that the proposed algorithm is learning a dependency between the context and the right exploration value, which is not the case of the min approach. This results show that when we are facing an non-stationary environment, having a dynamic exploration exploitation trade-off is useful, to learn this context/exploration dependency, which is beneficial for the contextual bandit algorithm.   

\begin{table}[H]
\centering
\label{tab:exp_1}
\begin{tabular}{l|lll}
switch iteration  & 100     & 1000   & 10 000      \\ \hline
max                       &  16123 & 15970    & 11063    \\
min                       &  15869  & \textbf{15331}    & \textbf{9291}    \\
mean                      &   15973.04 & 15584.21 & 10179.37  \\
median                    &  15960.5  & 15570    & 10200     \\
OPLINUCB              &  16061  & 15630    &    12601   \\
DOPLINUCB           &  13720  & \textbf{13569}    & 10702   

\end{tabular}
\caption{Performance measures for when reward function of each arm changes for scenario 2}
\end{table}

\section{Conclusion}
We have studied the problem of learning the exploration exploitation trad-off in the contextual bandit problem with linear reward function setting. In the traditional algorithms that solve the contextual bandit problem, the exploration is a parameter that is tuned by the user. However, our proposed algorithm learn to choose the right exploration parameters in an online manner based on the observed context, and the immediate reward received for the chosen action. We have presented here two algorithms that uses a bandit to find the optimal exploration of the contextual bandit algorithm. The Evaluation showed that the two proposed algorithms gives better results then  which we hope is the first step toward an automated multi-armed bandit algorithm. 


\bibliographystyle{aaai}
\bibliography{nips_2016}
\end{document}